\documentclass{article}
\usepackage{ijcai21}

\usepackage{times}
\usepackage{soul}
\usepackage{url}
\usepackage[hidelinks]{hyperref}
\usepackage[utf8]{inputenc}
\usepackage[small]{caption}
\usepackage{graphicx}
\usepackage{amsmath}
\usepackage{amsthm}
\usepackage{booktabs}
\usepackage{color}
\usepackage{makecell}
\usepackage{bigstrut}
\usepackage{multirow}
\usepackage{algorithmic}
\usepackage[linesnumbered,ruled]{algorithm2e}
\title{Learning from Self-Discrepancy via Multiple Co-teaching for Cross-Domain Person Re-Identification}

\author{
Suncheng Xiang$^{1}$\thanks{The corresponding author.}
\and
Yuzhuo Fu$^1$\and
Mengyuan Guan$^{1}$\And
Ting Liu$^1$
\affiliations
$^1$School of Electronic Information and Electrical Engineering\\
Shanghai Jiao Tong University, Shanghai, China
\emails
\{xiangsuncheng17, yzfu, gemini.my, louisa\_liu\}@sjtu.edu.cn
}

\begin{document}

\maketitle

\begin{abstract}
  Employing clustering strategy to assign unlabeled target images with pseudo labels has become a trend for person re-identification (re-ID) algorithms in domain adaptation. A potential limitation of these clustering-based methods is that they always tend to introduce noisy labels, which will undoubtedly hamper the performance of our re-ID system. To handle this limitation,
  an intuitive solution is to utilize collaborative training to purify the pseudo label quality. However, there exists a challenge that the complementarity of two networks, which inevitably share a high similarity, becomes weakened gradually as training process goes on; worse still, these approaches typically ignore to consider the self-discrepancy of intra-class relations. To address this issue, in this paper, we propose a multiple co-teaching framework for domain adaptive person re-ID, opening up a promising direction about self-discrepancy problem under unsupervised condition. On top of that, a mean-teaching mechanism is leveraged to enlarge the difference and discover more complementary features. Comprehensive experiments conducted on several large-scale datasets show that our method achieves competitive performance compared with the state-of-the-arts.
\end{abstract}

\section{Introduction}

\label{sec1}
Given a query image, person re-identification (re-ID) aims to match the person-of-interest across multiple non-overlapped cameras distributed in different places. Encouraged by the remarkable success of deep learning methods and the availability of large-scale datasets, re-ID research community has achieved significant progress during the past few years~\cite{zheng2016person,ye2021deep}. However, as for pedestrian images from an unseen domain, even with a large diversity of training data, person re-ID model generally experiences catastrophic performance drops because of the huge domain gaps or scene shifts, which cannot satisfy the need of application in real scenarios. To alleviate this problem, unsupervised domain adaptation (UDA)~\cite{ganin2015unsupervised,xiang2020unsupervised,saito2018open} is therefore proposed to employ the model trained on source dataset with identity labels to perform inference on the target domain. Nevertheless, it still remains an open research challenge in industry and academia due to the lack of identity annotations.

\begin{figure}[t]
\centering
\includegraphics[width=1.0\columnwidth]{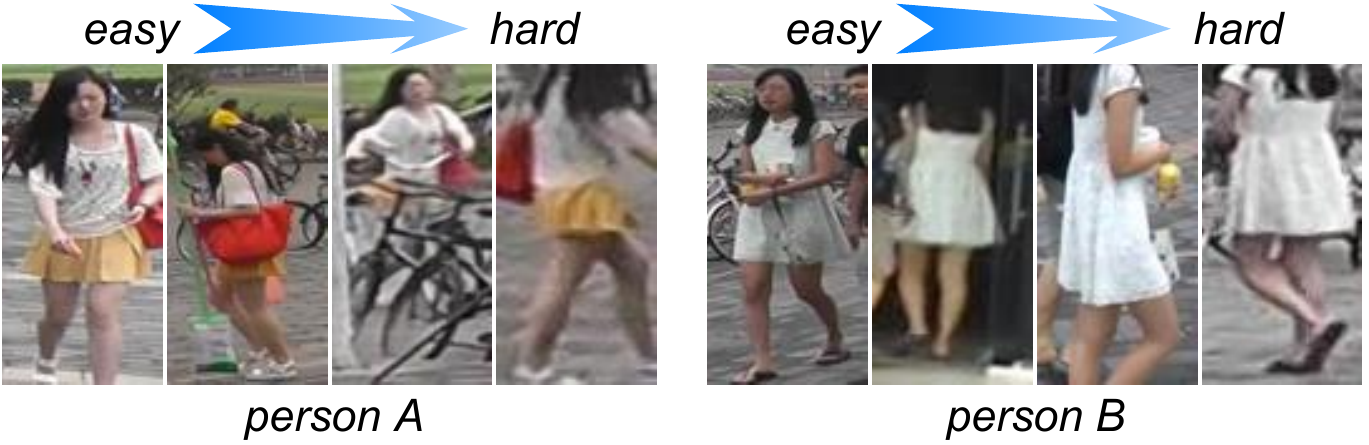}
\caption{Illustration of the self-discrepancy of intra-class relations for UDA person re-ID tasks, which is caused by variations in pose, viewpoint and occlusion, etc. For each identity, some \emph{\textbf{easy}} samples can be assigned with reliable pseudo labels. However, most of \emph{\textbf{hard}} samples are always given with noisy pseudo labels.}
\label{fig1}
\end{figure}

Currently, there are two main categories of UDA methods in re-ID community. The first category of image-level adaptation aims to eliminate the data distribution discrepancy across source and target domain, such as PTGAN~\cite{wei2018person} and SPGAN~\cite{deng2018image}. Although these approaches achieve promising progress, their performance deeply relies on the images generation quality. The second category of clustering-based adaptation~\cite{song2020unsupervised,fu2019self,fan2018unsupervised} deploys clustering algorithm to generate pseudo-labels for unsupervised target images during training period. Unfortunately, their abilities are substantially hindered by the inevitable label noises caused by imperfect clustering algorithms. To alleviate this problem, some co-teaching based re-ID approaches~\cite{yang2020asymmetric,ge2020mutual,zhao2020unsupervised,zhai2020multiple} have been introduced for combating with noisy labels after clustering.
Even though their optimal performance is often achieved by sub-network's discrimination ability, the self-discrepancy of intra-class relation (as shown in Figure~\ref{fig1}) in target domain still remains unexplored. So a natural
question then comes to our attention: \textit{how to leverage self-discrepancy features of multiple sub-network, and then optically adapt them to unlabelled domain,} which has to be fully elaborated.
Another challenge we observe is that, as the training process goes on, two neural networks in traditional co-teaching~\cite{han2018co} tend to converge and unavoidably share a high similarity, which weakens their complementarity and further improvement in terms of performance.

To solve the challenges mentioned above, we propose a simple yet powerful \textbf{M}ultiple \textbf{C}o-teaching \textbf{N}etwork \textbf{MCN} that considerably explores the self-discrepancy of intra-class relation in target domain, consequently, person re-ID can be more effectively performed to resist with noisy labels in domain adaptation. In addition, we introduce a mean-teaching mechanism to greatly enhance the complementarity and independence of collaborative networks, which, in turn, further improves the discriminability of learned representations in a progressive fashion. To the best of our knowledge, this is the first research effort to exploit the potential of \emph{self-discrepancy} among intra-class to address the UDA problem.  Compared with existing co-teaching based method~\cite{ge2020mutual,zhao2020unsupervised,zhai2020multiple}, our MCN is different from them in terms of \textbf{data input} and \textbf{model structure}: \textbf{(1)} Our work proposes to adopt samples with different discrepancy granularity ($T_{1}$ $\sim$ $T_{n}$) as asymmetric inputs to multiple networks, while previous methods applied same dataset as symmetric inputs during training; \textbf{(2)} MEB-Net~\cite{zhai2020multiple} used DenseNet-121~\cite{huang2017densely}, ResNet-50~\cite{he2016deep} and Inception-v3~\cite{szegedy2016rethinking} as backbone for enhancing the independence and complementary, \cite{ge2020mutual,zhao2020unsupervised} utilized random erasing or
random seeds for creating a difference, \cite{ge2020mutual,zhai2020multiple} also adopted symmetrical architecture with soft pseudo labels as well as hard pseudo labels in UDA re-ID tasks. In contrast, our MCN is only trained based on ResNet-50 with hard pseudo labels, which makes it more flexible and adaptable. In addition, our method can significantly mine the self-discrepancy feature in target domain, and a novel mean-teaching mechanism is also adopted to enhance the independence and complementary between teacher network and student networks, while previous asymmetric co-teaching approach~\cite{yang2020asymmetric} fails to meet these needs.


In total, our contribution can be summarized as follows:

1. We propose a multiple co-teaching network MCN to mine the self-discrepancy of intra-class relations
in target domain for solving noisy labels.

2. A \textbf{M}ean-\textbf{T}eaching mechanism is introduced to further enhance the output complementarity in a progressive manner based on proposed MCN method (``MCN-MT" for short).

3. Experimental results conducted on several benchmarks demonstrate the effectiveness of our proposed method.

\begin{figure*}[!t]
\centering{\includegraphics[width=1.0\linewidth]{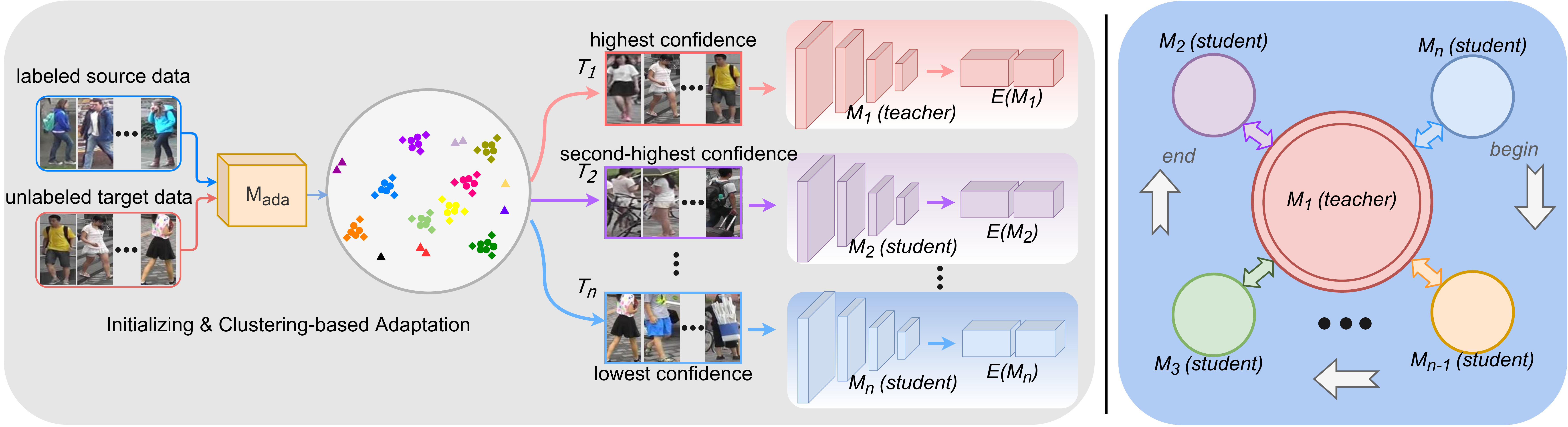}}
\caption{Overall framework of our multiple co-teaching method. $\mathcal{T}_{1}$ $\sim$ $\mathcal{T}_{n}$ denote target samples in different granularity levels of clustering confidences (highest $\rightarrow$ lowest). $E\left[M_{1}\right]$ $\sim$ $E\left[M_{n}\right]$ represent temporal average models of
$M_{1}$ $\sim$ $M_{n}$, which indicate multiple co-teaching networks corresponding to $\mathcal{T}_{1}$ $\sim$ $\mathcal{T}_{n}$, respectively. In the co-teaching paradigm, multiple student networks are organized to learn from teacher network in a progressive fashion ($M_{n}$ $\rightarrow$ $M_{2}$).}
\label{fig2}
\end{figure*}

\section{Related works}
\label{sec2}
\subsection{Unsupervised domain adaptation}
UDA aims to generalize the model learned from labeled source domain to the other unlabeled target domain, and the labeled and unlabeled examples are from non-overlapping classes. To build the learning relationship between them, many methods have been extensively explored in recent years, these works mainly lie in two aspect: image-level adaptation and clustering-based adaptation. The first one attempts to align the
source and target distribution. For instance, SPGAN~\cite{deng2018image} learns a similarity preserving GAN model by using the negative pairs to improve the image-image translation performance. Wei et al.~\cite{wei2018person} proposed a Person Transfer GAN network to bridge the domain gap between different styles of two datasets and migrate pedestrian style from one dataset to another. Although these unsupervised domain adaptation approaches achieve promising progress, their performance is still unsatisfactory compared with the fully supervised approaches~\cite{xiang2020multi}.
The latter attempts to estimate pseudo identity labels on the target domain so as to learn deep models in a supervised manner. For example, Fan et al.~\cite{fan2018unsupervised} propose a progressive unsupervised learning method consisting of clustering and fine-tuning the network. A self-training augmentation method PAST~\cite{zhang2019self} is proposed to promote the performance on target dataset progressively.
However, pseudo labels assigned by clusters can be very noisy as clustering
accuracy on hard samples is not satisfactory. Importantly, these methods either ignore to explore the self-discrepancy of intra-class relation in target domain, or suffer from the limitation of sub-optimal clustering results, which is not practical in real-world scenarios.

\subsection{Learning with noisy labels}
Deep learning with noisy labels is practically challenging, which has been widely studied in recent years. One of popular deep learning paradigm is co-teaching~\cite{han2018co}.
During the co-teaching,
the key idea of teacher-student models is to create consistent training supervision for labeled or unlabeled data via different models’ predictions,
then allow teacher and student networks to teach other in mutual perspective. Recently, this idea has been applied to distill powerful and easy-to-train large networks into small but harder-to-train networks~\cite{romero2014fitnets} that can even outperform their teacher. However,
there exists a problem which is ignored by majority of researchers:
the output of teacher network and student network might converge to equal each other and
the two networks tend to loss their output independence progressively when inputs of different branches share a high similarity or repeat with each others. Importantly,
existing teacher-student models could not be directly utilized on unsupervised domain adaptation (UDA) tasks of person re-ID since they are mostly designed for close-set recognition problems, which hinders the further improvement of UDA person re-ID task.

Aiming to address these challenges mentioned above, in this work, we creatively develop a multiple co-teaching network MCN to considerably
explore the self-discrepancy of intra-class relations in target domain, and then combat with noisy labels in domain adaptation. Furthermore,
in order to enhance the complementarity and avoid error amplification of different collaborative networks, we introduce a mean-teaching mechanism to boost the performance for UDA task prominently. This is the first time as far as we know, to comprehensively explore the self-discrepancy in target domain for re-ID task.


\section{Our Approach}
\subsection{Preliminary}
In the UDA re-ID task, we are given a labeled source dataset $\mathcal{S}=\left\{x_{1}, x_{2}, \cdots, x_{N}\right\}$, consisting
of $N_{s}$ person images with manually annotated labels $\mathcal{Y}=\left\{y_{1}, y_{2}, \cdots, y_{N}\right\}$. We also have a unlabeled target dataset $\mathcal{T}=\left\{t_{1}, t_{2}, \cdots, t_{M}\right\}$. Note that there is non-overlapping in terms of identity between source domain and target domain in open set domain adaptation. Our goal is to learn a feature embedding function that
can be applied to test set $\mathcal{X}^{t}=\left\{x_{1}^{t}, x_{2}^{t}, \ldots x_{N_{t}}^{t}\right\}$ of $N_{t}$ person images  and query set $\mathcal{X}^{q}=\left\{x_{1}^{q}, x_{2}^{q}, \ldots x_{N_{q}}^{q}\right\}$ of $N_{q}$ person images during the evaluation. By leveraging both labeled source images and unlabeled target images, we can learn a discriminative CNN model for UDA re-ID task.

\subsection{Multiple Co-teaching Network (MCN)}
\label{sec2.1}
To learn self-discrepancy of intra-class relations, we propose a multiple co-teaching network MCN which trains several networks progressively with samples at different granularity levels. As shown in Figure~\ref{fig2}, MCN consists one teacher network $M_{1}$ and several student networks $M_{2}$ $\sim$ $M_{n}$. Specifically, we firstly train CNN on the source labeled data and fine-tune it on target data with pseudo labels to get initial weights for $M_{1}$ $\sim$ $M_{n}$, then we perform multiple co-teaching paradigms between \emph{\textbf{teacher}} network and several \emph{\textbf{student}} networks. In particular, teacher network receives highest confidence samples ($\mathcal{T}_{1}$) as much as possible while student networks take in samples with lower confidence levels ($\mathcal{T}_{n}$, $\mathcal{T}_{n-1}$, $\cdots$,$\mathcal{T}_{2}$) as diverse as possible. In the first paradigm of co-teaching, \emph{\textbf{teacher}} network $M_{1}$ performs co-teaching with \emph{\textbf{student}} network $M_{n}$ until they reach convergence, followed by the second co-teaching paradigm between teacher network $M_{1}$ and student network $M_{n-1}$.
Note that there are \text{n-1} co-teaching paradigms in total needed to be performed between teacher network $M_{1}$ and several student networks $M_{2}$ $\sim$ $M_{n}$, respectively. To be more specific, the student networks select diversified samples from lower confidence set to train teacher network when multiple students are involved progressively, which encourages the teacher network to have a basic discriminability for representation learning.

\subsection{Mean-Teaching Mechanism}
\label{sec2.2}
In traditional co-teaching, a popular strategy is to employ the predictions of teacher model for training other student networks. However, most of researchers always neglect that directly using the current predictions to train student models degrades the complementary of teacher models' outputs~\cite{tarvainen2017mean}. To address this issue, we introduce a mean-teaching mechanism to greatly enhance the independence and complementary of teacher network and student networks. To be more specific, mean-teaching leverages the temporally average models of networks to generate pseudo labels for supervising each other. During the training iteration $T$, the parameters of the temporally average models are denoted as $\Theta\left(E^{T}\left[M\right]\right)$, which can be calculated as
\begin{equation}
\Theta\left(E^{T}\left[M\right]\right)=\alpha * \Theta\left(E^{T-1}\left[M\right]\right)+(1-\alpha) * \Theta \left(M\right)
\label{eq1}
\end{equation}
where $\Theta\left(E^{T-1}\left[M\right]\right)$ indicates the temporal average parameters of the networks in the previous iteration $T$-1, the initial temporal average parameters are $\Theta\left(E^{0}\left[M\right]\right) = \Theta \left(M\right)$, $\alpha$ is the hyper-parameter within the range [0,1). Different from~\cite{tarvainen2017mean} whose weight is temporal average of the student network parameters, our teacher model is trained with diverse samples mined by student networks, which encourages the teacher network to receive samples as diverse as possible, so the weights of MCN-MT can be dynamic updated as training goes on.
The pseudo hard labels of both average model and its peer network are utilized jointly to train the several collaborative networks. During the evaluation period, we adopt the past average model of teacher network for down-stream re-ID task.

\begin{algorithm}[t]
     \caption{The training procedure of our method}
     \small
     \label{alg1}
      \KwIn{Labeled source dataset $\mathcal{S}$, unlabeled target dataset $\mathcal{T}$,
       CNN model $M$, granularity level $n$ of self-discrepancy, maximum iteration round $r$.}
      \KwOut{Best model $M_{best}$ \& $E^{T+1}\left[M_{best}\right]$.}
        \tcp{***baseline initialization***}
        $M_{src}$ $\gets$ Initialize $M$ on $\mathcal{S}$\; \label{code1}
        \tcp{***clustering-based adaptation***}
        Divide $\mathcal{T}$ into inliers $\mathcal{T}_{in}$ and outliers $\mathcal{T}_{out}$ by DBSCAN clustering results\; \label{code2}
        $\mathcal{T}_{n}$ $\gets$ $\mathcal{T}_{out}$; k = 1 \;
        \Repeat{n = k + 1}
        {
            Divide $\mathcal{T}_{in}$ into inliers $\mathcal{T}_{in}^{k}$ and outliers $\mathcal{T}_{out}^{k}$ by DBSCAN  clustering results \; \label{code2}
            $\mathcal{T}_{in}$ $\gets$ $\mathcal{T}_{in}^{k}$, $\mathcal{T}_{n-k}$ $\gets$ $\mathcal{T}_{out}^{k}$; k ++ \;
        }
        $\mathcal{T}_{1}$ $\gets$ $\mathcal{T}_{in}$ \;
        $M_{ada}$ $\gets$ Fine-tune $M_{src}$ with $\mathcal{T}_{1}$ $\cup$ $\mathcal{T}_{2}$ $\cup$ $\cdots$ $\cup$ $\mathcal{T}_{n-1}$  \; \label{code3}
        $M_{1}$ $\gets$ $M_{ada}$, $M_{2}$ $\gets$ $M_{ada}$, $\cdots$, $M_{n}$ $\gets$ $M_{ada}$ \; \label{code4}
         \tcp{***multiple co-teaching***}
        \For{i = n $\to$ 2}
        {
            \For{T = 1 $\to$ r}
            {
                \eIf{T \% 2 == 0}
                    {
                        Deploy $M_{i}$ to select reliable instances from $\mathcal{T}_{i}$ for optimizing $M_{1}$, then update $E^{T+1}\left[M_{1}\right]$ \; \label{code5}
                        }{
                        Deploy $M_{1}$ to select reliable instances from $\mathcal{T}_{1}$ for optimizing $M_{i}$, then update $E^{T+1}\left[M_{i}\right]$ \; \label{code6}
                         }
            }
        }
        \textbf{Return} best model $M_{best}$ \& $E^{T+1}\left[M_{best}\right]$ \;
 \end{algorithm}

\subsection{Dynamic Network Updating}
As shown in the Algorithm~\ref{alg1}, we firstly use pre-trained ResNet-50~\cite{he2016deep} on ImageNet~\cite{deng2009imagenet} for initializing with source dataset $\mathcal{S}$, then adopt source model $M_{src}$ to extract pooling-5 features of target images $\mathcal{T}$, which assigns reliable pseudo hard labels for exemplars in high-density area while noisy pseudo labels for samples in low-density area. We set a hyper-parameter \emph{n} to control the granularity of discrepancy and a hyper-parameter \emph{r} to represent maximum iteration round during training. Consequently, target images $\mathcal{T}$ can be divided into \emph{n} granularity levels ($\mathcal{T}_{1}$ $\sim$ $\mathcal{T}_{n}$ sets) based on the clustering results.
In particular, $M_{ada}$ is fine-tuned over diverse set $\mathcal{T}_{1}$, $\mathcal{T}_{2}$, ..., $\mathcal{T}_{n}$, which acts as a warm start for training multiple student networks, then we perform several co-teaching paradigms between teaching network and student networks progressively. In this paper, the noisy pseudo labels caused by clustering, which
result in a decline in performance, can be alleviated by our MCN framework with mean-teaching induction, this gives rise to our MCN-MT method.

During the training, we use triplet loss~\cite{hermans2017defense} to mine the relationship of training samples, which can minimize the distance among positive pairs and maximize the distance between negative pairs. And our loss is defined as:
\begin{equation}
\mathcal{L}_{triplet}=\left(d_{a, p}-d_{a, n}+m\right)_{+}
\label{eq2}
\end{equation}
where $d_{a, p}$, $d_{a, n}$ denote the feature distances of positive pair and negative pars, respectively, $m$ represents the margin of our triplet loss, $(z)_{+}$ denotes \textit{max(z,0)}.

\section{Experiment}
\subsection{Datasets}
We conduct experiments on three benchmark datasets, including Market-1501~\cite{zheng2015scalable}, DukeMTMC-reID~\cite{ristani2016performance,zheng2017unlabeled} and CUHK03~\cite{li2014deepreid}. Market-1501 has 1,501 identities in 32,668 images. 12,936 images of 751 identities are used for training, the query has 3,368 images and gallery has 19,732 images. DukeMTMC-reID contains 16,522 images of 702 identities for training, and the remaining images of 702 identities for testing. CUHK03 consists of 14,097 images with a total 1,467 identities.
We evaluate the quality of our model using mean average precision (mAP) and Cumulative Matching Characteristic (CMC) curves.

\subsection{Implementation Details}
In this paper, we follow the training procedure in~\cite{yang2020asymmetric} and
empirically set $\alpha = 0.999$ in Eq.~\ref{eq1}. The batch size of training samples is set as 64.
As for triplet selection, we randomly selected 16 persons and sampled 4 images for each identity, $m$ is set as 0.5 in Eq.~\ref{eq2}.
Meanwhile, all experiments is based on DBSCAN clustering~\cite{ester1996density}, the minimum size of a cluster is constrained to 4 and the density radius is set to $1.6 \times 10^{-3}$. In multiple co-teaching paradigm, we set maximum iteration rounds r = 30 until it reaches convergence state.
\begin{table}[t]
  \centering
  \caption{Ablation study. We evaluate the performance (\%) of our proposed MCN and the simple fine-tuning respectively. ``Direct transfer" means a model trained on source dataset is directly adopted for evaluation on target dataset.}
  \setlength{\tabcolsep}{2.4mm}{
    \begin{tabular}{lcccc}
    \toprule
    \multirow{2}[4]{*}{Method} & \multicolumn{2}{c}{Duke $\rightarrow$ Market} & \multicolumn{2}{c}{Market $\rightarrow$ Duke} \\
\cmidrule{2-5}          & R-1 & mAP   & R-1 & mAP \\
    \midrule
    Direct transfer & 57.6  & 20.6  & 28.3  & 15.2  \\
    Fine-tuning  & 72.8  & 51.6  & 64.3  & 46.8  \\
    MCN (Ours)  & 82.6  & 63.2  & 72.5  & 53.5  \\
    \bottomrule
    \end{tabular}}%
  \label{tab3}%
\end{table}%


\subsection{Ablation Study}
To further validate the effectiveness of the our proposed method, we perform several ablation studies on the individual component of our proposed multiple co-teaching method.

\textbf{The effectiveness of proposed MCN:}
To argue the effectiveness of our proposed method MCN, we conduct extensive experiments under another setting of simple fine-tuning with single network.
As depicted in Table~\ref{tab3}, it can be easily observed that our multiple co-teaching network can achieve more competitive performance than fine-tuning with target images in cross-domain re-ID task, \textit{e.g.}, our MCN method can achieve 82.6\% in rank-1 accuracy and 63.2\% in mAP on Market-1501, however, it can only obtain 72.8\% in rank-1 accuracy and 51.6\% in mAP with simple fine-tuning strategy. Not surprisingly, mAP accuracy is also significantly reduced from 53.5\% to 46.8\% if directly applying fine-tuning on DukeMTMC-reID benchmark, which demonstrates the superiority of our proposed MCN method.

\begin{figure}[!t]
\begin{minipage}[t]{0.5\linewidth}
\centering
\includegraphics[width=1.70in]{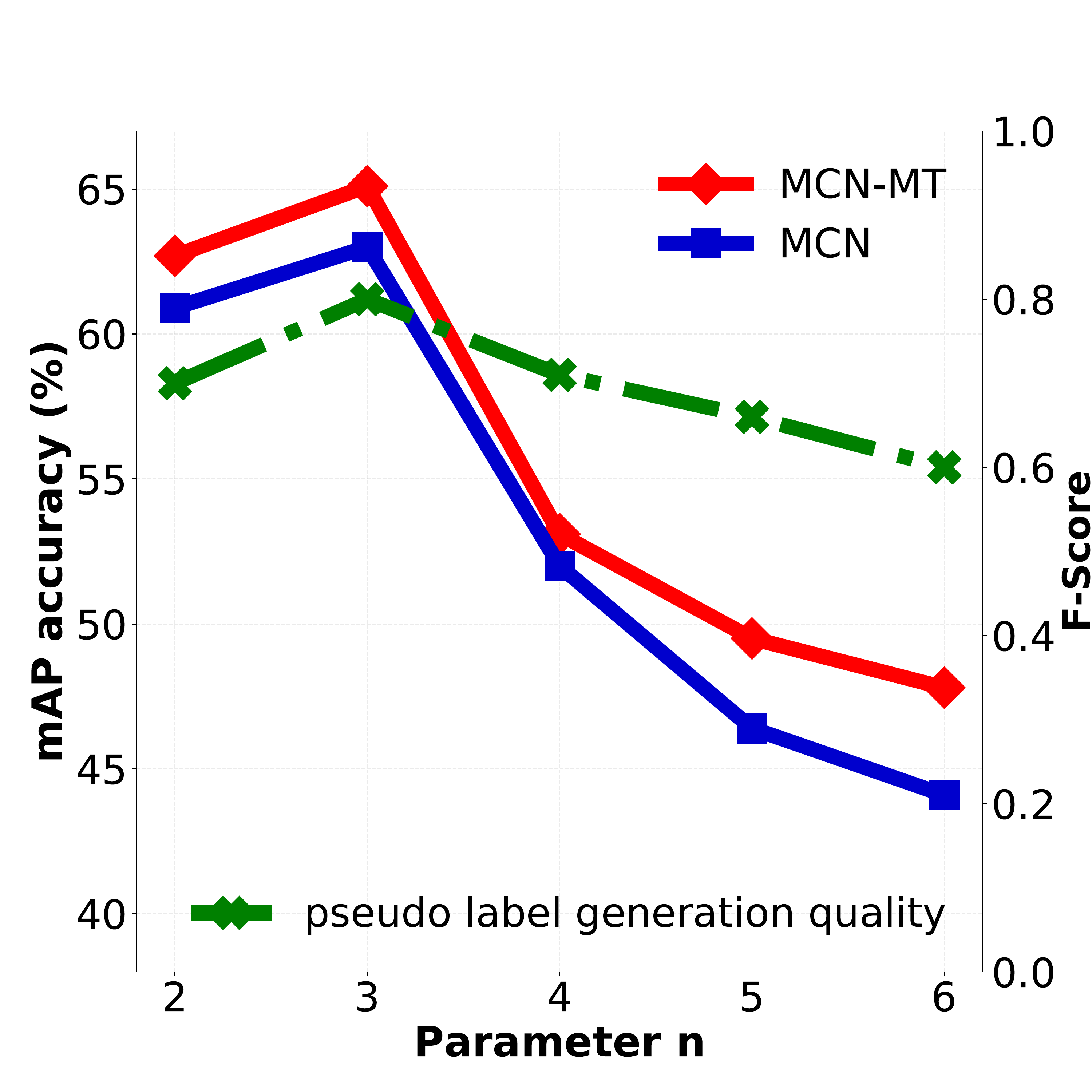}
\centerline{(a) Duke$\rightarrow$Market}
\label{fig51}
\end{minipage}%
\begin{minipage}[t]{0.5\linewidth}
\centering
\includegraphics[width=1.70in]{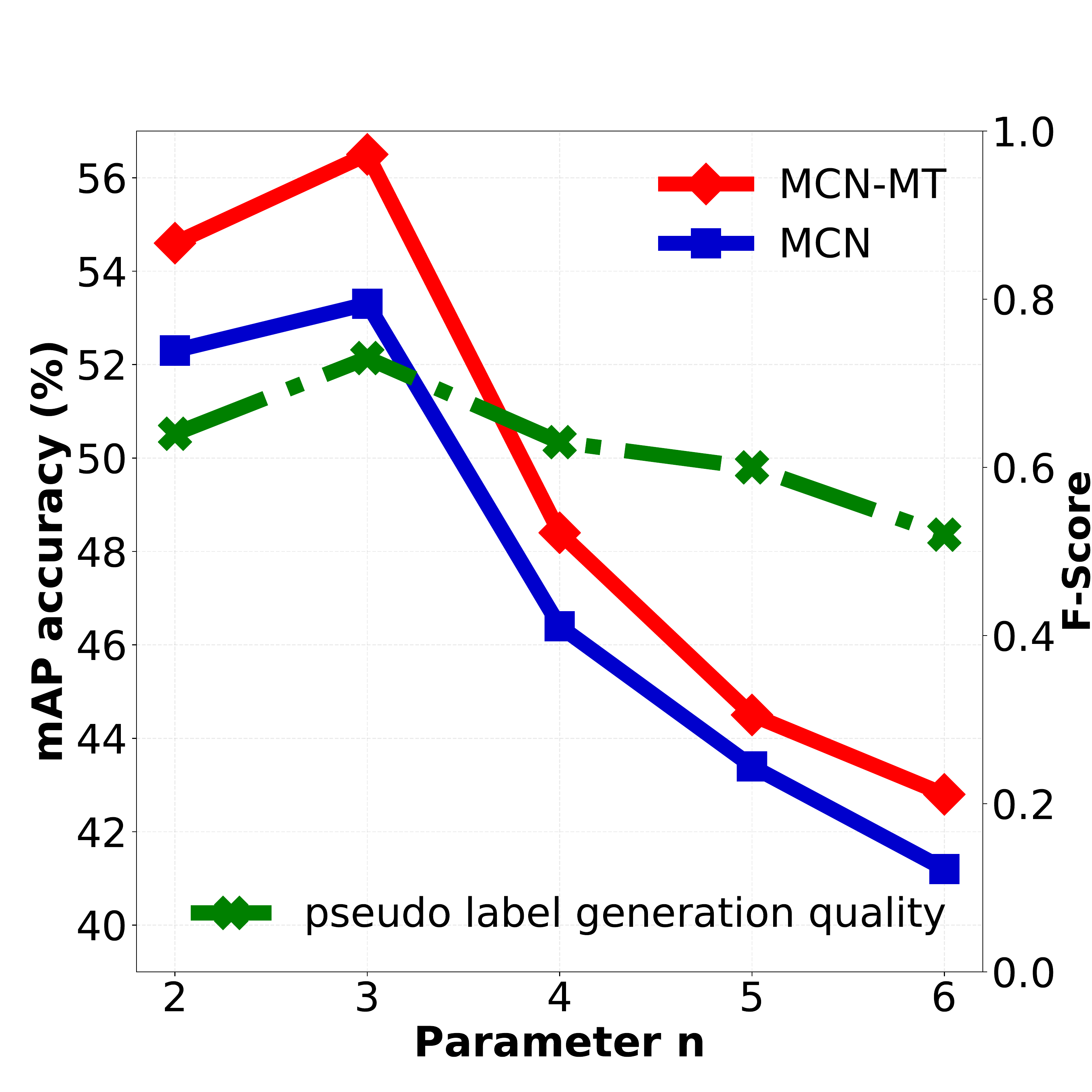}
\centerline{(b) Market$\rightarrow$Duke}
\label{fig52}
\end{minipage}
 \caption{Impact of $n$. mAP accuracy are compared. We adopt average F-score of MCN-MT (higher is better; marked in \textcolor[rgb]{0,0.4,0}{green} dash-line) to measure the quality of pseudo label generation on Market and Duke, respectively.}
\label{fig5}
\end{figure}

\textbf{The effectiveness of mean-teaching mechanism:}
We evaluate the mean-teaching component proposed in Section~\ref{sec2.2}. As illustrated in Table~\ref{tab1} and Figure~\ref{fig5}, when n=3, results show that mAP drops from 64.9\% to 63.2\% on Market-1501 and from 57.8\% to 53.5\% on DukeMTMC-reID without adopting mean-teaching induction. Similar drops can also be observed no matter which self-discrepancy parameter is employed in MCN. The effectiveness of the mean-teaching can be largely attributed to that it enhances the discrimination capability of all collaborative networks during multiple co-teaching.

\textbf{The impact of discrepancy granularity:}
Intuitively, $n$ determines the granularity of the self-discrepancy relations, when $n$=2, only two models are trained collaboratively. As $n$ increases, retrieval accuracy improves at first. However, mAP accuracy does not always increase with confidence granularity level $n$. As illustrated in Figure~\ref{fig5}, when $n$=5 or 6, the performance drops dramatically, regardless of using mean-teaching induction. To go even further, we gave an explanation about this phenomenon from two aspects: \textit{Qualitative perspective} and \textit{Quantitative perspective}.

\begin{figure}[!t]
\centering{\includegraphics[width=0.9\linewidth]{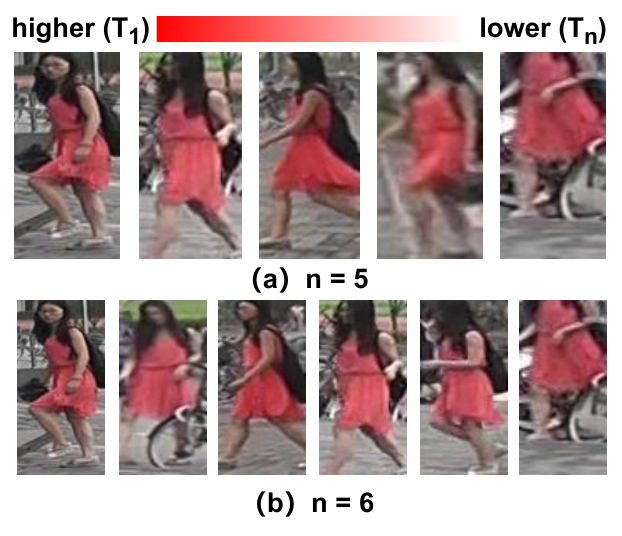}}
\caption{Visualization of the self-discrepancy under different $n$ values. When \textbf{(a)} $n$=5 or \textbf{(b)} $n$=6, some samples of same pedestrian in different self-discrepancy granularity repeat with others or share a high similarity.}
\label{fig6}
\end{figure}

First, from the qualitative perspective,
a visualization of the self-discrepancy offers explanations to this phenomenon, as shown in Fig.~\ref{fig6}. When $n$ equals to 5 or 6, some images of same pedestrian in different confidences are very similar to each other, so the inputs of different branches share a high similarity,
which may degrades the complementary capacity of MCN-MT.

\begin{table*}[t]
  \centering
  \caption{Performance (\%) comparisons with the state-of-the-art methods. \textcolor{red}{\textbf{Red}} indicates the best and \textcolor{blue}{\textbf{Blue}} the second best.}
  \setlength{\tabcolsep}{2.9mm}{
    \begin{tabular}{lcccccccc}
    \toprule
    \multirow{2}[4]{*}{Method} & \multicolumn{4}{c}{DukeMTMC-reID$\rightarrow$Market-1501} & \multicolumn{4}{c}{Market-1501$\rightarrow$DukeMTMC-reID} \\
\cmidrule{2-9}          & R-1 & R-5  & R-10  & mAP   & R-1 & R-5  & R-10  & mAP \\
    \midrule
    PTGAN~\cite{wei2018person} & 38.6  & 57.3  & 66.1  & 15.7  & 27.4  & 43.6  & 50.7  & 13.5  \\
    SPGAN~\cite{deng2018image} & 51.5  & 70.1  & 76.8  & 22.8  & 41.1  & 56.6  & 63.0  & 22.3  \\
    HHL~\cite{zhong2018generalizing} & 62.2  & 78.8  & 84.0  & 31.4  & 46.9  & 61.0  & 66.7  & 27.2  \\
    ECN~\cite{zhong2019invariance} & 75.1  & 87.6  & 91.6  & 43.0  & 63.3  & 75.8  &  80.4  & 40.4  \\
    MAR~\cite{yu2019unsupervised} & 67.7  & 81.9  & -  & 40.0  & 67.1  & 79.8  & -  & 48.0  \\
    SSG~\cite{fu2019self} & 80.0  & 90.0  & 92.4  & 58.3  & 73.0  & 80.6  & 83.2  & 53.4  \\
    PCB-PAST~\cite{zhang2019self} & 78.4  & -  & -  & 54.6  & 72.4  & -  & -  & 54.3  \\
    ACT~\cite{yang2020asymmetric} & 80.5  & -  & -  & 60.6  & 72.4  & -  & -  & 54.5  \\
    MMT~\cite{ge2020mutual} (\textit{w/} $\mathcal{L}_{stri}^{t}$)   & \textcolor{blue}{\textbf{84.0}}  & \textcolor{blue}{\textbf{93.4}}  & \textcolor{blue}{\textbf{95.4}}  & \textcolor{blue}{\textbf{62.6}}  & \textcolor{red}{\textbf{74.9}}  & \textcolor{red}{\textbf{85.2}}  & \textcolor{red}{\textbf{89.5}}  & \textcolor{red}{\textbf{58.1}}  \\
    \midrule
    MCN (Ours)   & 82.6  & 90.7  & 94.1  & 63.2  & 72.5  & 81.8  & 84.6  & 53.5  \\
    MCN-MT (Ours)   & \textcolor{red}{\textbf{84.3}}  & \textcolor{red}{\textbf{93.6}}  & \textcolor{red}{\textbf{95.9}}  & \textcolor{red}{\textbf{64.9}}  & \textcolor{blue}{\textbf{74.7}}  & \textcolor{blue}{\textbf{83.8}}  & \textcolor{blue}{\textbf{86.3}}  & \textcolor{blue}{\textbf{57.8}}  \\
    \bottomrule
    \end{tabular}}%
  \label{tab1}%
\end{table*}

\begin{table*}[t]
  \centering
  \caption{Performance (\%) comparisons with the state-of-the-art methods. \textcolor{red}{\textbf{Red}} indicates the best and \textcolor{blue}{\textbf{Blue}} the second best.}
  \setlength{\tabcolsep}{3.8mm}{
    \begin{tabular}{lcccccccc}
    \toprule
    \multirow{2}[4]{*}{Method} & \multicolumn{4}{c}{CUHK03$\rightarrow$Market-1501} & \multicolumn{4}{c}{CUHK03$\rightarrow$DukeMTMC-reID} \\
\cmidrule{2-9}          & R-1 & R-5  & R-10  & mAP   & R-1 & R-5  & R-10  & mAP \\
    \midrule
    PTGAN~\cite{wei2018person} & 31.5  & -  & 60.2  & -   & 17.6  & -  & 38.5  & -  \\
    SPGAN~\cite{deng2018image} & 42.3  & -  & -  & 19.0  & -  & -  & -  & - \\
    HHL~\cite{zhong2018generalizing} & 56.8  & 74.7  & 81.4  & 29.8  & 42.7  & 57.5  & 64.2  & 23.4  \\
    EANet~\cite{huang2018eanet} & 66.4  & -  & -  & 40.6  & 45.0  & -  & -  & 26.4  \\
    ACT~\cite{yang2020asymmetric} & \textcolor{blue}{\textbf{81.2}}  & -  & -  & \textcolor{blue}{\textbf{64.1}}  & \textcolor{blue}{\textbf{52.8}}  & -  & -  & \textcolor{blue}{\textbf{35.4}}  \\
    \midrule
    MCN (Ours)  & 82.2  & 92.4  & 95.5  & 66.1  & 53.3  & 66.3  & 71.3  & 37.2  \\
    MCN-MT (Ours)  & \textcolor{red}{\textbf{84.8}}  & \textcolor{red}{\textbf{93.1}}  & \textcolor{red}{\textbf{95.7}}  & \textcolor{red}{\textbf{68.7}}  & \textcolor{red}{\textbf{56.3}}  & \textcolor{red}{\textbf{67.3}}  & \textcolor{red}{\textbf{73.3}}  & \textcolor{red}{\textbf{40.2}}  \\
    \bottomrule
    \end{tabular}}%
  \label{tab2}%
\end{table*}%

Second, from the quantitative perspective,
our qualitative observations above are confirmed by the quantitative evaluations. To be more specific, we use F-score to measure the pseudo label generation quality of our proposed MCN-MT method. As depicted in Fig.~\ref{fig5}, the quality of pseudo labels will be negatively affected with an over-increased n. when $n$ equals to 3, we can obtain the highest F-score on Market-1501 and DukeMTMC-reID datasets respectively, suggesting the high quality of our pseudo labels, which may be the main reason that the proposed MCN-MT can achieve the best performance when $n$=3.
In real-world applications, we would recommend to use $n$=3.


\begin{table}[t]
  \centering
  \caption{Ablation study. We evaluate the performance (\%) of our proposed MCN-MT with hard pseudo labels and soft pseudo labels respectively.}
  \setlength{\tabcolsep}{2.0mm}{
    \begin{tabular}{lcccc}
    \toprule
    \multirow{2}[4]{*}{Method} & \multicolumn{2}{c}{Duke $\rightarrow$ Market} & \multicolumn{2}{c}{Market $\rightarrow$ Duke} \\
\cmidrule{2-5}          & R-1 & mAP   & R-1 & mAP \\
    \midrule
    MCN-MT (soft)  & 77.8  & 59.8  & 71.2  & 52.7  \\
    MCN-MT (hard)  & 84.3  & 64.9  & 74.7  & 57.8  \\
    \bottomrule
    \end{tabular}}%
  \label{tab4}%
\end{table}%

\subsection{Comparison with the State-of-the-art Methods}
\textbf{Market-1501:}
From Table~\ref{tab1} and Table~\ref{tab2}, it can be seen clearly that our MCN-MT (\textit{w/} Mean-Teaching) achieves remarkable rank-1 accuracy of 84.3\% and 84.8\% when trained on DukeMTMC-reID and CUHK03 respectively, outperforming the second-best methods MMT~\cite{ge2020mutual} and ACT~\cite{yang2020asymmetric} by \textbf{+2.3\%} and \textbf{+4.6\%} in mAP accuracy. The superiority of our proposed method can be largely contributed to
the self-discrepancy of intra-class relations mined by MCN-MT during multiple collaborative training, which is beneficial to learn a more robust and discriminative model in UDA re-ID tasks.

\textbf{DukeMTMC-reID:}
When performing evaluation on DukeMTMC-reID dataset, our approach has also achieved superior results than state-of-the-art methods on this dataset, which is the most challenging dataset currently with some occlusion. Comparing to MMT~\cite{ge2020mutual}, our model obtains nearly similar mAP score when trained on Market-1501, but achieving a higher rank-1 score leading by \textbf{+3.5\%} comparing to ACT~\cite{yang2020asymmetric} when trained on CUHK03. It is worth noting that MMT utilizes soft softmax-triplet loss with soft triplet labels, and performance of MMT-500 (\textit{w/} $\mathcal{L}_{stri}^{t}$) \& (\textit{w/o} $\mathcal{L}_{sid}^{t}$) is reported in Table~\ref{tab1}, which indicates that MMT is more complex than our proposed method and this may be the main reason leading to the better performance when trained on Market-1501.

\subsection{Discussion}
As shown in Table~\ref{tab1}, when tested on Market-1501$\rightarrow$DukeMTMC-reID, we find an interesting phenomenon that performance
of MCN-MT is slightly inferior and less competitive compared with MMT~\cite{ge2020mutual} (\textit{w/} $L_{stri}^{t}$).
Generally speaking, soft pseudo-labels perform relatively better than hard labels in computer vision tasks with
symmetric networks. In essence, we have indeed performed some experiments by
adopting soft pseudo-labels in our MCN-MT method, which is generated by the past
temporally average model of teacher and student networks. However, we found that our
model is really hard to reach a convergence state with soft pseudo-labels during training
process, which leads to significantly performance degradation on DukeMTMC-reID
and Market-1501 dataset. As shown in Table~\ref{tab4}, e.g., with soft pseudo labels, we can only achieve a mAP
accuracy of 59.8\% on DukeMTMC-reID$\rightarrow$Market-1501, and 52.7\% on Market-1501$\rightarrow$DukeMTMC-reID respectively. We
suspect this is due to the inputs of multiple student networks MCN-MT are asymmetric
and have a large difference in terms of self-discrepancy relations. As a result, the soft
pseudo-labels of same images generated by the past temporally average model cannot
maintain their consistency, which undoubtedly has negative impacts on the training of
multiple co-teaching networks. This motivates us to perform further research of multiple co-teaching strategy for domain adaptation in the future.

\section{Conclusion and Future Work}
\label{sec4}
In this paper, we firstly present a simple yet effective multiple co-teaching network MCN to mine the self-discrepancy in target domain for UDA re-ID task, which trains several neural networks simultaneously with unlabeled samples in coarse-grained discrepancy. Furthermore, a novel mean-teaching induction is introduced to further enlarge the difference and learn discriminative features on the basis of MCN. By plugging our mean-teaching mechanism into MCN, the complementarity of the teacher network and student network is significantly enhanced. Comprehensive experiments conducted on benchmark datasets show that our method outperforms the state-of-the-art UDA methods by a clear margin. As a future direction, we
will extend our method to handle with other challenging computer vision tasks, such as fine-grained image retrieval.

\section{Acknowledgments}
\label{sec5}
This research was partially supported by the National Natural Science Foundation of China under grant No. 61977045.

\bibliographystyle{named}
\bibliography{ijcai21}

\end{document}